\documentclass[10pt, a4paper]{article}
\usepackage{lrec2022} 
\usepackage{multibib}
\newcites{languageresource}{Language Resources}
\usepackage{graphicx}
\usepackage{tabularx}
\usepackage{soul}

\usepackage{titlesec}
\titleformat{\section}{\normalfont\large\bfseries\center}{\thesection.}{1em}{}
\titleformat{\subsection}{\normalfont\SmallTitleFont\bfseries\raggedright}{\thesubsection.}{1em}{}
\titleformat{\subsubsection}{\normalfont\normalsize\bfseries\raggedright}{\thesubsubsection.}{1em}{}
\renewcommand\thesection{\arabic{section}}
\renewcommand\thesubsection{\thesection.\arabic{subsection}}
\renewcommand\thesubsubsection{\thesubsection.\arabic{subsubsection}}

\usepackage{epstopdf}
\usepackage[utf8]{inputenc}

\usepackage{hyperref}
\usepackage{xstring}
\usepackage{color}

\usepackage{float}

\title{The TalkMoves Dataset: K-12 Mathematics Lesson Transcripts Annotated for Teacher and Student Discursive Moves}

\name{\parbox{\linewidth}{\centering Abhijit Suresh$^{1,2}$, Jennifer Jacobs$^{2}$, Charis Harty$^{2}$, Margaret Perkoff$^{1}$, \\ James H. Martin$^{1,2}$, Tamara Sumner$^{1,2}$}}

\address{ \\
         $^1$Department of Computer Science, $^2$Institute of Cognitive Science\\
         University of Colorado Boulder \\
         Firstname.Lastname@colorado.edu \\
         }

\abstract{
Transcripts of teaching episodes can be effective tools to understand discourse patterns in classroom instruction. According to most educational experts, sustained classroom discourse is a critical component of equitable, engaging, and rich learning environments for students. This paper describes the TalkMoves dataset, composed of 567 human-annotated K-12 mathematics lesson transcripts (including entire lessons or portions of lessons) derived from video recordings. The set of transcripts primarily includes in-person lessons with whole-class discussions and/or small group work, as well as some online lessons. All of the transcripts are human-transcribed, segmented by the speaker (teacher or student), and annotated at the sentence level for ten discursive moves based on accountable talk theory. In addition, the transcripts include utterance-level information in the form of dialogue act labels based on the Switchboard Dialog Act Corpus. The dataset can be used by educators, policymakers, and researchers to understand the nature of teacher and student discourse in K-12 math classrooms. Portions of this dataset have been used to develop the TalkMoves application, which provides teachers with automated, immediate, and actionable feedback about their mathematics instruction.
 \\ \newline \Keywords{Classroom transcripts, K-12 Education, Accountable Talk, Dialog Acts} }

\begin{document}

\maketitleabstract

\section{Introduction}

Recordings of classroom activities - including video, audio, and transcripts - provide essential data sources for understanding contemporary classroom dynamics and pedagogies \cite{major2018using}\cite{xu2018video}, and for training new forms of intelligent educational technologies. Unfortunately, due to practical and privacy concerns, classroom recordings often have limited availability and sharing is very restricted \cite{derry2010conducting}. This lack of access to valuable resources and shared data sets has hindered the field's ability to replicate and build on published work. Here, we release and introduce a new corpus:  the TalkMoves dataset of over 500  K-12 mathematics lesson transcripts enriched with annotations that capture important, research-based aspects of classroom discourse.   

There is widespread agreement that student understanding is strongly enhanced through sustained interaction within a learning community and that content-rich discussions should be a prominent and normative feature within K-12 classrooms \cite{franke2015student}.  In an effort to delineate specific practices that teachers can use to orchestrate the types of discussions encouraged by the Common Core State Standards for Mathematical Practice \cite{national2010common}, researchers developed an approach to classroom discourse called “accountable talk” \cite{o2015scaling}. At the heart of accountable talk is the notion that teachers should organize discussions that promote students’ equitable participation in a rigorous learning environment where their thinking is made explicit and publicly available to everyone in the classroom. 

Research that aims to support positive changes in teaching and learning should build on curated resources that are based on explicit and evidence-based theories of learning. The TalkMoves dataset contains annotations that are firmly grounded in a coherent educational theory, specifically accountable talk, that is supported by a large body of empirical research on teaching and learning. As such, this dataset offers a unique collection of textual materials that are well-suited for natural language processing (NLP) and machine learning applications.

\section{Related Work}
\subsection{Accountable Talk Theory}

Accountable talk theory identifies and defines an explicit set of discursive techniques that can promote rich, knowledge-building discussions in classrooms. These well-defined discursive techniques have been incorporated into a wide range of instructional practices and frameworks (e.g., \cite{michaels2010accountable};\cite{boston2012assessing}; \cite{candela2020discourse}), and their specificity and definitions make them ripe for natural language processing. 

These discursive techniques center on “talk moves”, which refer to specific dialog acts intended to elicit a response by another member of the class \cite{o2019supporting}. Both teachers and learners can use talk moves to construct conversations in which students share their thinking, actively consider the ideas of others, and engage in sustained reasoning. Teacher talk moves include questions that press students to justify their thinking or to assess the contributions made by other students, positioning students as capable mathematics learners  \cite{michaels2010accountable}. Student talk moves are discourse actions such as making claims, using reasoning, reacting to other students’ ideas, and asking questions. By using these moves, students contribute their ideas and attend to and build on their classmates' ideas, helping to ensure they are actively and equitably engaged in challenging academic work \cite{candela2020discourse}.

Within the accountable talk framework, talk moves are clustered into three categories based on their instructional purpose \cite{resnick2018accountable}: (1) accountability to the learning community, (2) accountability to content knowledge, and (3) accountability to rigorous thinking. The released version of the TalkMoves dataset and the currently deployed TalkMoves application (talkmoves.com) include six teacher and four student talk moves, drawn from all three categories. Table \ref{tab:talkmoves} provides a brief description of each teacher and student talk move, along with illustrative examples. 

These ten talk moves were selected for inclusion due to their relatively high frequency in our transcript corpus, the ability of human coders to establish high interrater reliability, and guidance from experts in accountable talk \cite{jacobs2021}. This set of talk moves is not exhaustive; there are other important talk moves, including those that have (and potentially have not yet) been identified and labeled as such in the research literature \cite{o2019supporting}. 

\subsection{Related Corpora}

There are several existing conversation datasets that include annotations for dialogue acts. The most prominent of these is the DA-labeled Switchboard corpus  \cite{stolcke2000dialogue}  which contains conversation transcripts between pairs of participants. Similarly, the ICSI meetings dataset includes recordings and annotated transcripts from  75 meetings \cite{janin2003icsi}. This dataset comprises dialog act annotations in the Meeting Recorder Dialog Act corpus \cite{shriberg2004icsi}. More recently, the MultiWOZ dataset was introduced as a multi-domain conversational dataset \cite{budzianowski2018multiwoz} and the EMOTyDA dataset \cite{saha2020towards} was introduced, which contains multi-modal data with annotations for dialog acts as well emotions. 

While there is a growing set of conversational resources that researchers can build on, none of these datasets capture teaching and learning conversations that are unique to the classroom environment. Classroom discourse differs from other conversational contexts due to a number of inherent characteristics: it often involves a large number of unique and overlapping speakers, including both children or youth and adults. One of the primary bottlenecks for developing computational models of classroom discourse is a lack of publicly available data. The closest domain-relevant dataset to the Talkmoves dataset is the Educational Uptake dataset which provides annotations for math classroom transcripts that include uptake indicators to signify teacher engagement \cite{demszky2021measuring}.

\begin{table*}
\centering
\caption{Teacher and student talk moves included in the TalkMoves dataset and application}
\label{tab:talkmoves}
\begin{tabular}{|p{1.8cm}|p{2.3cm}|p{4.5cm}|p{4.5cm}|}
\hline
\multicolumn{1}{|c|}{\textbf{Category}} & \multicolumn{1}{c|}{\textbf{Talk move}}       & \multicolumn{1}{c|}{\textbf{Description}}                                                                                    & \multicolumn{1}{c|}{\textbf{Example}}                                      \\ \hline
\multicolumn{4}{|c|}{Teacher Talk Moves}                                                                                                                                                                                                                                                            \\ \hline
Learning Community                      & Keeping everyone together                     & Prompting students to be active listeners and orienting students to each other                                               & “What did Eliza just say her equation was?”                                \\ \hline
Learning Community                      & Getting students to relate to another’s ideas & Prompting students to react to what a classmate said                                                                         & “Do you agree with Juan that the answer is 7/10?”                          \\ \hline
Learning Community                      & Restating                                     & Repeating all or part of what a student said word for word                                                                   & “Add two here.”                                                            \\ \hline
Content Knowledge                       & Pressing for accuracy                         & Prompting students to make a mathematical contribution or use mathematical language                                          & “Can you give an example of an ordered pair?”                              \\ \hline
Rigorous Thinking                       & Revoicing                                     & Repeating what a student said but adding on or changing the wording                                                          & “Julia told us she would add two here.”                                    \\ \hline
Rigorous Thinking                       & Pressing for reasoning                        & Prompting students to explain, provide evidence, share their thinking behind a decision, or connect ideas or representations & “Why could I argue that the slope should be increasing?”                   \\ \hline
\multicolumn{4}{|c|}{Student Talk Moves}                                                                                                                                                                                                                                                            \\ \hline
Learning Community                      & Relating to another student                   & Using, commenting on, or asking questions about  a classmate’s ideas                                                         & “I didn’t get the same answer as her.”                                     \\ \hline
Learning Community                      & Asking for more info                          & Student requests more info, says they are confused or need help                                                              & “I don’t understand number four.”                                          \\ \hline
Content Knowledge                       & Making a claim                                & Student makes a math claim, factual statement, or lists a step in their answer                                               & “X is the number of cars.”                                                 \\ \hline
Rigorous Thinking                       & Providing evidence or reasoning               & Student explains their thinking, provides evidence, or talks about their reasoning                                           & “You can’t subtract 7 because then you would only get 28 and you need 29.” \\ \hline
\end{tabular}
\end{table*}

\section{Dataset Description}

The TalkMoves dataset can be used to generate meaningful insights into language-centered approaches to teacher and student learning, student engagement, and structures and participation in knowledge-building conversations. Our released TalkMoves dataset includes 567 transcripts, comprising  174,186 annotated teacher utterances, 59,874 student utterances, and 1.8 million words (15,830 unique). Each transcript generally encompasses an entire mathematics lesson (typically 55 minutes long), but occasionally transcripts for short excerpts from larger lessons are also included. In all cases, the transcripts were human-generated from classroom audio and/or video recordings. All of the transcripts in the dataset were human-annotated for the 10 teacher and student talk moves listed in Table \ref{tab:talkmoves}. In addition to the human annotation of accountable talk moves, all transcripts in the TalkMoves dataset have also been coded with computationally derived dialogue acts (DAs) \cite{jurafsky1997switchboard}. A detailed coding protocol supporting these 10 talk moves is included in this dataset. 

\subsection{Data sources for the TalkMoves Dataset}
This dataset is derived from three pre-existing public collections of transcripts as well as anonymized transcripts collected as part of the TalkMoves project. The pre-existing transcripts were drawn from the following sources: Inside Mathematics (educational resources compiled by the Charles A. Dana Center at the University of Texas at Austin, https://www.insidemathematics.org), the Third International Mathematics and Science Study (TIMSS) 1999 video study (public use resources collected as part of the study, http://www.timssvideo.com), and Video Mosaic (educational resources curated by the Robert B. Davis Institute for Learning at Rutgers University, https://videomosaic.org). Additional transcripts were collected through the online TalkMoves application (talkmoves.com) described later in this paper (collectively called “The TalkBack studies”). These pre-existing sites were selected as data sources as they are all recognized providers of research-based mathematics education resources, and the free use and distribution of the provided recording or transcripts is supported. Members of the research team who are experts in mathematics education reviewed each recording or transcript prior to inclusion in the data set.

\subsection{TalkMoves Annotation}

All transcripts in the dataset were human-annotated for 10 teacher and student talk moves. Annotations are applied at the sentence level; i.e., each teacher and student sentence is “tagged” to indicate which type of talk move (including “none”) it is representative of.  A sample annotated excerpt can be found in Table \ref{tab:sample}. 

\begin{table*}
\centering
\caption{Sample annotated excerpt from a classroom session}
\label{tab:sample}
\begin{tabular}{|p{1cm}|p{5cm}|p{2cm}|p{2cm}|}
\hline
\multicolumn{1}{|c|}{\textbf{Speaker}} & \multicolumn{1}{c|}{\textbf{Sentence}}                                                                          & \multicolumn{1}{c|}{\textbf{Teacher Tag}} & \multicolumn{1}{c|}{\textbf{Student Tag}}     \\ \hline
Teacher                                & Look, we have a  different model over here, even.                                                               & 1 - None                                  &                                               \\ \hline
Teacher                                & So now we have three.                                                                                           & 1 - None                                  &                                               \\ \hline
Teacher                                & I wonder if  it’s going to be the same as yours, or if it’s going to be the same as  this one.                  & 1 - None                                  &                                               \\ \hline
Teacher                                & Is two thirds still bigger, Greg, is two thirds still bigger than  a half, on this model too, or did it change? & 8 - Press for Accuracy                    &                                               \\ \hline
Teacher                                & Ok, Danielle, what do you think about this time?                                                                & 8 - Press for Accuracy                    &                                               \\ \hline
Student                                & Well, um, two thirds                                                                                            &                                           & 4 - Making a Claim                            \\ \hline
Teacher                                & What is two thirds?                                                                                             & 8 - Press for Accuracy                    &                                               \\ \hline
Teacher                                & Can you build a two thirds and a one half for  him separate so we can then compare?                             & 8 - Press for Accuracy                    &                                               \\ \hline
Student                               & Here’s the two thirds, and here’s the half                                                                      &                                           & 4 - Making a Claim                            \\ \hline
Teacher                                & What’s the difference?                                                                                          & 8 - Press for Accuracy                    &                                               \\ \hline
Student                                & and it’s bigger by two twelfths.                                                                                &                                           & 5 - Providing Evidence / Explaining Reasoning \\ \hline
Student                               & It’s,  um, it’s bigger by two twelfths                                                                          &                                           & 5 - Providing Evidence / Explaining Reasoning \\ \hline
Teacher                                & Oh, so is he getting a different answer from that, too, or are they the  same?                                  & 3 - Getting Students to Relate            &                                               \\ \hline
\end{tabular}
\end{table*}

\subsubsection{Gold Standard Reliability}

The team worked with experts in math education and accountable talk to develop a detailed coding protocol. Two members of the TalkBack research team served as annotators. These annotators established an initial inter-rater agreement using the protocol before applying the talk moves codes to the corpus. They also calculated their agreement when they were approximately halfway through coding to ensure that their annotations remained accurate and consistent. Their reliability, calculated using Cohen’s kappa \cite{mchugh2012interrater}, was high for each talk move at both periods (see Table \ref{tab:reliability}). Such high reliability among human experts is critical for ensuring that machine learning models will be able to accurately discriminate between these different labels.

\begin{table*}[ht]
\centering
\caption{ Reliability scores for each teacher and student talk move}
\label{tab:reliability}
\begin{tabular}{|p{5cm}| p{2cm}| p{2cm}| p{2cm}|} 
 \hline
\bf Coding decision & \bf Inter-rater agreement & \bf Initial kappa & \bf Midpoint kappa  \\ [0.5ex] 
 \hline
 Keeping everyone together & 88\% & 0.91 & 0.96\\
 \hline
 Getting students to relate & 94\% & 0.91 & 0.92\\
 \hline
  Restating & 100\% & 1.0 & 1.0 \\
 \hline
  Revoicing & 98\%  & 0.99 & 1.0\\
 \hline
  Press for accuracy & 89\% & 0.93 & 0.95\\
 \hline
  Press for reasoning & 92\% & 0.95& 0.95 \\ [1ex] 
 \hline
\end{tabular}
\end{table*}

\subsubsection{Data Preprocessing}

The transcripts were pre-processed through multiple steps to prepare them for human annotation and model development. First, each raw transcript was converted into a Comma Separated Values (CSV) file using an automated script. Because the transcripts were obtained from multiple sources, they initially had different formatting conventions and layouts, which we then standardized. Finally, we removed metadata introduced during the transcription process  (eg. “[background noise]”). 

The converted CSV files include six columns: Time-stamp, Turn, Speaker, Sentence, Teacher Tag and Student Tag. The number of rows is equivalent to the number of sentences in the transcript. The Timestamp variable indicates the beginning time for each sentence, if available. Sentences that are spoken by the same speaker without interruption are considered part of the same turn. Turns are numbered sequentially (from 1-n) throughout each transcript. The Speaker variable identifies the sentence as spoken by the teacher or a student. When teachers or individual students were named in the original transcript, these names are included in the Speaker column. All proper names have been anonymized in the transcripts collected for the TalkBack studies. Finally, the Teacher and Student tag refer to the annotated teacher and student talk moves, respectively. Six mutually exclusive teacher talk moves (or “none”) were tagged for each sentence spoken by a teacher. Four mutually exclusive student talk moves (or “none”) were tagged for each sentence spoken by a student.

\subsubsection{Uneven Distribution of Talk Moves in the Dataset}

Of note is the uneven distribution pattern of the talk moves included in the TalkMoves dataset, with certain talk moves being much more frequently used during classroom lessons than others (see Table \ref{tab:talkmovesdist}). This distribution pattern reflects natural variation in how teachers and students use talk moves in mathematics lessons, with some moves being more common than others. Furthermore, talk moves are “special” linguistic acts, meaning that when they occur they have a particular meaning for both the speaker and the listeners. Therefore it is not surprising that among all of the teacher and student sentences, the most common talk move label is “none,” indicating that those sentences do not contain a talk move. For teacher sentences that have a talk move, the two most common moves are Keeping Everyone Together and Pressing for Accuracy. The most common student talk move is Making a Claim, which typically co-occurs with teacher’s Pressing for Accuracy. The skewed nature of this type of real-world data presents classification challenges that remain unresolved in the field of machine learning and deep learning \cite{krawczyk2016learning}.

\begin{table}[ht]
\centering
\caption{Distribution of teacher and student talkmoves }
\label{tab:talkmovesdist}
\begin{tabular}{|p{3.5cm}| p{3cm}|} 
 \hline
\bf Teacher TalkMove & \bf  \% utterances in TalkMoves dataset  \\ [0.5ex] 
 \hline
 Keeping everyone together & 13.075\% \\
 \hline
 Getting students to relate & 1.643\% \\
 \hline
  Restating & 1.5\%  \\
 \hline
  Revoicing & 2.295\% \\
 \hline
  Press for accuracy & 13.161\% \\
 \hline
  Press for reasoning & 1.17\% \\  
 \hline
  No TalkMove & 67.154\% \\
 \hline
 \bf Student TalkMove & \bf  \% utterances in TalkMoves dataset  \\ [0.5ex]
 \hline
 Relating to another student & 11.108\% \\
 \hline
 Asking for more info & 3.203\% \\
 \hline
 Making a claim & 30.624\% \\
 \hline
 Providing evidence & 13.353\%  \\
 \hline
  No TalkMove & 41.71\% \\[1ex]
 \hline
\end{tabular}
\end{table}

\subsection{Dialog Act Annotations}

In addition to the human annotation of accountable talk moves, the transcripts in the TalkMoves dataset have also been coded with computationally derived dialogue acts (DAs). Dialogue acts are labels that provide sentence-level pragmatic information.  This type of information is beneficial for modeling the overall flow of a conversation between one or more individuals. Furthermore, dialog acts can be used as an additional semantic feature for multi-task models in order to supplement word vectors. Ideally, this will allow us to improve the accuracy of our Talk Moves classification model in future work. For the TalkMoves dataset, we adopted the Switchboard Dialog Act Corpus (SWBD-DAMSL) framework, which is composed of 42 DA labels \cite{jurafsky1997switchboard}. The DA label corresponding to each utterance was calculated using a self-governing neural network (github.com/glicerico/SGNN) based on the work of Ravi and Kozareva \cite{ravi2018self}. Although the DAs have not yet been included as features in the deep learning models used in the TalkMoves application, it is possible that doing so may further improve their performance in the automated identification of teacher and student talk moves. 

\subsubsection{Distribution of Dialog Acts}

Among the 42 possible DA labels, only seven unique DA labels appear in the transcripts in the TalkMoves dataset (see Table \ref{tab:dialog}). Perhaps to be expected for mathematics lessons, Statement-Non-Opinion was the most prominent label.  This label is also the most frequent for the original Switchboard corpus, so it is possible that the high prevalence of it in the TalkMoves dataset is due to the fact that the model is trained on the Switchboard dataset. Utterances with this code can take a number of forms.  Some samples in the dataset include ``I hear some wonderful thinking here." and ``This was not an easy one."  For future work, it may be beneficial to fine tune the self-governing neural network on the TalkMoves dataset after collecting human annotated dialog act codes.  Additionally, it may make sense to use a condensed version of the SWBD-DAMSL tag set that is more relevant in the classroom context.

\begin{table}[ht]
\centering
\caption{Distribution of Dialog Act labels }
\label{tab:dialog}
\begin{tabular}{|p{3.8cm}| p{3.1cm}|} 
 \hline
\bf Dialog Act label & \bf  \% utterances in TalkMoves dataset  \\  
 \hline
Ackonwledge(Backchannel) & 10.954\% \\
 \hline
 Agree/Accept & 3.233\% \\
 \hline
  Appreciation & 2.125\%  \\
 \hline
  Yes-No-Question & 0.457\% \\
 \hline
  Uninterpretable & 1.778\% \\
 \hline
  Conventional closing & 0.004\% \\  
 \hline
  Statement opinion & 4.258\% \\
 \hline
  Statement non-opinion & 77.189\% \\ 
 \hline
\end{tabular}
\end{table}

\subsection{The TalkMoves application}

A challenge of critical importance within education is providing teachers with timely and detailed feedback about their classroom discourse. Currently, such feedback is only sporadically provided as it requires highly trained classroom observers, and it is time-consuming and expensive to deploy such observers in classrooms. The TalkMoves application was designed to automate and scale up the process of detecting and classifying talk moves, along with other classroom discourse practices, enabling teachers to receive immediate and accessible information about their mathematics lessons. The application consists of three interrelated components: a cloud-based big data infrastructure for managing and processing classroom recordings, deep learning models that reliably detect the use of teacher and student talk moves, and an interface that provides teachers with personalized feedback on their use of discussion strategies \cite{suresh2018using}, \cite{suresh2021using}.

TalkMoves offers an example of an NLP application that supports a well-specified theory of learning (accountable talk), addresses a recognized challenge in education (teacher feedback), and potentially scales to large numbers of teachers. This effort demonstrates how a new form of big data - classroom recordings - can be leveraged with advances in automated speech recognition and deep learning models to provide teachers with unique insights into their instruction. Initial evidence suggests that this information is perceived as valuable and actionable by teachers \cite{karla2021} and increases teachers’ use of talk moves over time \cite{jacobs2021}.

\begin{figure*}[ht]
\center
  \includegraphics[scale=.45]{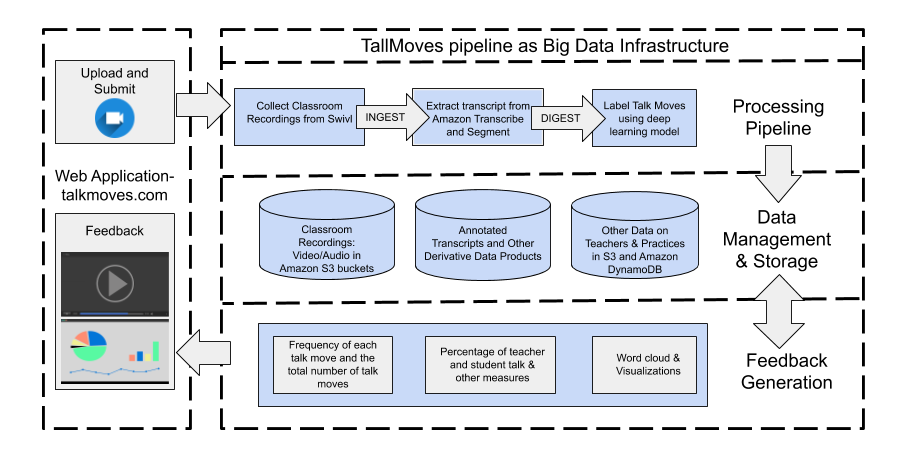}
  \caption{System architecture}
  \label{fig:1}
\end{figure*}

The system architecture of the TalkMoves application includes a processing pipeline, data management and storage, and feedback generation (Figure \ref{fig:1}) \cite{jacobs2021}. First, teachers generate and upload classroom recordings, which can consist of entire lessons or portions of lessons. Next, the system collects the files, processing one video at a time through the pipeline. The audio is converted into a written transcript, which is then broken into sentences. Each sentence is designated as originating from the teacher or a student. Deep learning models then determine whether there is a talk move corresponding to each teacher or student sentence \cite{suresh2019automating}, \cite{suresh2021usingAI}. Additional analytics are applied to calculate other discursive features, such as how much talk came from the teacher versus the students. Finally, the system generates feedback based on the output from the model, which is visually displayed on a personalized dashboard using a web interface. For each uploaded recording the current interface displays the lesson video, a word cloud showing the most frequently used words, information about the teacher’s talk moves, information about the students’ talk moves, and additional discourse information (such as the percentage of teacher and student talk, wait time, one-word answers, and mathematical vocabulary). The interface also shows teachers how the data for a given lesson compare to their average (across all of their lessons) as well as the average across all of the current users’ lessons. Additionally, the interface includes resources about accountable talk theory, definitions and examples of each talk move, and how the application was developed. 

\subsection{Benchmarking the TalkMoves application}

To automate the identification of TalkMoves in classroom discourse and benchmark the performance on the publicly accessible dataset, we trained two separate models to identify the teacher and student talk moves. The transcripts were used for training and testing with a 90/10 split. A portion of the confidential dataset was used as the validation set for hyperparameter tuning. There is no overlap between the teachers in training and testing set to avoid the possibility of overfitting. We fine-tuned  transformers on the TalkMoves dataset for sequence classification \cite{devlin2018bert}.

For the teacher model, the inputs were student-teacher ``sentence pairs," which refers to a combination of a teacher sentence concatenated with the immediately prior student sentence. For example, a sentence pair can include a student utterance, ``I said, they hit their growth spurt earlier," followed by a teacher utterance, ``Okay, why do you think they hit their growth spurt earlier?". This sentence pair is a good example of the teacher encouraging the student to reason (pressing for reasoning). The output was a 7-way sequence classification (softmax) over the six teacher talk moves and ``None." 

Similarly, we applied transformers to classify student talk moves. The inputs to the student model were student-student ``sentence pairs," which refer to a combination of a student sentence concatenated with the immediately prior student sentence. An example student-student pair would be a student utterance, ``They are separated" followed by ``The histograms are all together." This exchange is a good example of a student making a claim. The output was a 5-way sequence classification (softmax) over the four student talk moves and ``None." The performance is measured based on macro-F1 scores and MCCs (Matthew Correlation Coefficient) on the testing set (as seen in Table \ref{tab:performance}). We did not find evidence for a significant change in performance on other variants of BERT \cite{liu2019roberta} \cite{clark2020electra}. 

For parameter selection, we considered the following variables: learning rate (2e-5, 3e-5, 4e-5, 5,e-5), number of epochs (3-6), batch size (4,8,16,32), warmup steps (0,100,1000) and maximum sequence length (128,256,512). We trained the models with an exhaustive choice of these parameters using Amazon EC2 instance (g3.16xlarge) with 4 Tesla M60 GPUs in parallel. The code was implemented in Python 3.7 with Pytorch and HuggingFace library. 

\begin{table}
\centering
\caption{Performance on test set with +- 0.5 error}
\label{tab:performance}
\begin{tabular}{|l|l|l|}
\hline & F1 score (in \%)  & MCC                                                     \\ \hline
\begin{tabular}[c]{@{}l@{}}\textbf{Teacher Model}\\ - BERT-base \\ - RoBERTa- base \\ - Electra-base \\ \end{tabular} & \begin{tabular}[c]{@{}l@{}}\\75.38\\ \textbf{76.32} \\ \textbf{75.77}\end{tabular} &
\begin{tabular}[c]{@{}l@{}}\\0.7438\\ \textbf{0.7513} \\ \textbf{0.7447}\end{tabular} \\ \hline
\begin{tabular}[c]{@{}l@{}}\textbf{Student Model}\\ - BERT-base \\ - RoBERTa-base \\ - Electra-base \\\end{tabular}  & \begin{tabular}[c]{@{}l@{}}\\\textbf{73.12}\\ 71.58 \\ 71.28\end{tabular} & \begin{tabular}[c]{@{}l@{}}\\\textbf{0.6716}\\ 0.6605 \\ 0.6447\end{tabular} \\ \hline
\end{tabular}
\end{table}

\section{Discussion}

\subsection{Data availability}

The TalkMoves application provides one example of how the TalkMoves dataset of lesson transcripts can be utilized to generate reliable deep learning models and incorporate them in a practical application for teachers. Other uses for the dataset are likely to be for similar linguistically motivated computational models that may have educational applications, such as models that look at talk moves in more nuanced ways (under different classroom conditions) and AI-based tools that “coach” teachers (and potentially students) as they engage in instructional activities. The dataset may also be applicable for natural language processing efforts that use computational models to explore classroom talk's nature, mechanics, and function. The TalkMoves dataset and accompanying resources are available in an online repository through GitHub (https://github.com/SumnerLab/TalkMoves). This repository includes:
\begin{enumerate}
    \item A spreadsheet listing all of the transcripts, including whether they are publicly accessible or confidential, and other descriptive information.
    \item Individual CSV files for each publicly accessible transcript, annotated for teacher, student talk moves, and dialog act labels based on the DAMSL framework.
    \item An annotation protocol describing each type of talk move, including definitions and examples.
\end{enumerate}

\subsection{Limitations}

In the U.S. alone, there are well over 100,000 schools serving students in grades K-12. Each of these schools likely offers numerous mathematics classes, often every day during the school year. It is nearly impossible to estimate the number of mathematics lessons that occur in the US during a given year, much less globally and over many years. The TalkMoves dataset incorporates transcripts from only a tiny fraction of these lessons. Nonetheless, recording and transcribing authentic mathematics classroom lessons is no simple matter. Generally, numerous permissions must be obtained, including voluntary informed consent from each child in the clasroom’s parent or guardian. Data that can be widely shared requires a much more concerted effort. Therefore, the TalkMoves dataset is likely to be highly valuable to researchers interested in a relatively large corpus of real-world classroom lesson transcripts, despite its apparent limitations in terms of scope and representation.

Most of the transcripts in the dataset were collected from lessons recorded in the U.S., although the TIMSS video study data does include classrooms filmed internationally. Where possible, descriptive information about each lesson is included, such as date recorded, teacher gender, grade level, original language, and whether the transcript is from a full or partial lesson. However, in many cases, some or all of this information is unknown. The available descriptive information indicates that the dataset is skewed towards female-taught U.S. middle school lessons (grades 6-8).

\subsection{Ethical Considerations}
The research team took into account a variety of important ethical considerations related to the release of the TalkMoves dataset. The full dataset includes publicly available transcripts on the internet at the time they were curated, transcripts provided to the research team by confidential sources, and transcripts collected by the research team as part of the TalkBack studies. The publicly available transcripts have not been altered in any way except as described in this paper, including steps undertaken for preprocessing and annotation of talk moves. The transcripts provided by confidential sources are not included in the set of available transcripts. Transcripts collected by the TalkBack research team are also described in relatively general terms to ensure that individual teachers and their students can not be identified. Additionally, all proper names in the TalkBack transcripts were anonymized prior to their inclusion in the dataset. The TalkBack studies were reviewed and approved by the University of Colorado Boulder's Institutional Review Board (IRB Protocol \#18-0432). 
\section{Conclusion}
Recently there has been an explosion in efforts to develop advanced algorithms in machine learning and natural language processing that can be applied to classroom data. This includes research efforts such as ours, work by \cite{suresh2021using} \cite{suresh2021usingAI}, and commercial products such as TeachFX (teachfx.com). Despite this growing interest in AI in education, there are very few publicly available datasets devoted to K-12 classroom data. To the best of our knowledge, the TalkMoves dataset is one of the few such meticulously curated datasets to be made available, and the only one that has been developed in accordance with a well-documented theoretical framework (accountable talk, \cite{michaels2010accountable}). 

This dataset is already providing numerous opportunities for researchers interested in the crossroads of natural language processing and education. For instance, one of the three National AI education-focused institutes funded by the National Science Foundation is already using this dataset. In a recent study, their researchers used accountable talk theory to develop the future talk move prediction (FTMP) task \cite{ganesh2021would}. FTMP can potentially be used to train conversational AI agents to provide immediate feedback to teachers on their classroom discourse. Other researchers have used this data set to study the relationship between accountable talk moves and dialog act labels; while others have examined the generalizability of the framework and models to other disciplines such as science classrooms. We have also shared this dataset with researchers developing child language models for automatic speech recognition systems tailored to school environments. Similar frameworks are also being used to understand conversational interaction between teachers and students \cite{demszky2021measuring}. To date, sharing this data has involved arduous processes involving multiple institutional review boards. Due to this clear demand, we prioritized making these data resources more broadly and publicly available to the research and development community.

\section{Acknowledgements}
The research team would like to thank Eddie Dombower and his team at Curve 10 for their contributions to the design and implementation of the TalkBack application. This material is based upon work supported by the National Science Foundation under Grant Numbers 1600325 and 1837986. This research was supported by the NSF National AI Institute for Student-AI Teaming (iSAT) under grant DRL 2019805. The opinions expressed are those of the authors and do not represent views of the NSF.
\vfill

\section{Bibliographical References}\label{reference}

\bibliographystyle{lrec2022-bib}
\bibliography{lrec2022-example}

\end{document}